\documentclass{article}

\usepackage{amsmath, amssymb}           
\usepackage{graphicx}                   
\usepackage{esvect}                     
\usepackage{hyperref}
\usepackage{pdfpages}
\usepackage{xcolor}
\usepackage{float}
\usepackage[sc]{titlesec}
\usepackage{natbib}
\usepackage{iclr2017_conference,times}
\usepackage{wrapfig}
\usepackage{cleveref}
\usepackage{dsfont}

\title{Gaussian Error Linear Units (GELUs)}

\author{Dan Hendrycks\thanks{Work done while the author was at TTIC. Code available at \href{https://github.com/hendrycks/GELUs}{github.com/hendrycks/GELUs} }\\University of California, Berkeley\\\texttt{hendrycks@berkeley.edu} \And Kevin Gimpel\\ Toyota Technological Institute at Chicago\\\texttt{kgimpel@ttic.edu}}

\begin{document}
\maketitle

\begin{abstract}
We propose the Gaussian Error Linear Unit (GELU), a high-performing neural network activation function. The GELU activation function is $x\Phi(x)$, where $\Phi(x)$ the standard Gaussian cumulative distribution function. The GELU nonlinearity weights inputs by their value, rather than gates inputs by their sign as in ReLUs ($x\mathbf{1}_{x>0}$). We perform an empirical evaluation of the GELU nonlinearity against the ReLU and ELU activations and find performance improvements across all considered computer vision, natural language processing, and speech tasks.
\end{abstract}

\section{Introduction}
Early artificial neurons utilized binary threshold units \citep{hopfield, mcculloch}. These hard binary decisions are smoothed with sigmoid activations, enabling a neuron to have a ``firing rate'' interpretation and to train with backpropagation. But as networks became deeper, training with sigmoid activations proved less effective than the non-smooth, less-probabilistic ReLU \citep{relu} which makes hard gating decisions based upon an input's sign. Despite having less of a statistical motivation, the ReLU remains a competitive engineering solution which often enables faster and better convergence than sigmoids. Building on the successes of ReLUs, a recent modification called ELUs \citep{elu} allows a ReLU-like nonlinearity to output negative values which sometimes increases training speed. In all, the activation choice has remained a necessary architecture decision for neural networks lest the network be a deep linear classifier.

Deep nonlinear classifiers can fit their data so well that network designers are often faced with the choice of including stochastic regularizer like adding noise to hidden layers or applying dropout \citep{dropout}, and this choice remains separate from the activation function. Some stochastic regularizers can make the network behave like an ensemble of networks, a pseudoensemble \citep{bachman}, and can lead to marked accuracy increases. For example, the stochastic regularizer dropout creates a pseudoensemble by randomly altering some activation decisions through zero multiplication. Nonlinearities and dropout thus determine a neuron's output together, yet the two innovations have remained distinct. More, neither subsumed the other because popular stochastic regularizers act irrespectively of the input and nonlinearities are aided by such regularizers.

In this work, we introduce a new nonlinearity, the Gaussian Error Linear Unit (GELU). It relates to stochastic regularizers in that it is the expectation of a modification to Adaptive Dropout \citep{standout}. This suggests a more probabilistic view of a neuron's output. 
We find that this novel nonlinearity matches or exceeds models with ReLUs or ELUs across tasks from computer vision, natural language processing, and automatic speech recognition.

\section{GELU Formulation}

We motivate our activation function by combining properties from dropout, zoneout, and ReLUs. First note that a ReLU and dropout 
both 
yield a neuron's output with the ReLU deterministically multiplying the input by zero or one and dropout stochastically multiplying by zero. Also, a new RNN regularizer called zoneout stochastically multiplies inputs by one \citep{zoneout}. We merge this functionality by multiplying the input by zero or one, but the values of this zero-one mask are stochastically determined 
while also 
dependent upon the input. Specifically, we can multiply the neuron input $x$ by $m \sim \text{Bernoulli}(\Phi(x))$, where $\Phi(x) = P(X\le x), X\sim \mathcal{N}(0,1)$ is the cumulative distribution function of the standard normal distribution. We choose this distribution since neuron inputs tend to follow a normal distribution, especially with Batch Normalization. In this setting, inputs have a higher probability of being ``dropped'' as $x$ decreases, so the transformation applied to $x$ is stochastic yet depends upon the input.
\begin{wrapfigure}{r}{0.58\textwidth}
	\centering
    \includegraphics[width=0.56\textwidth]{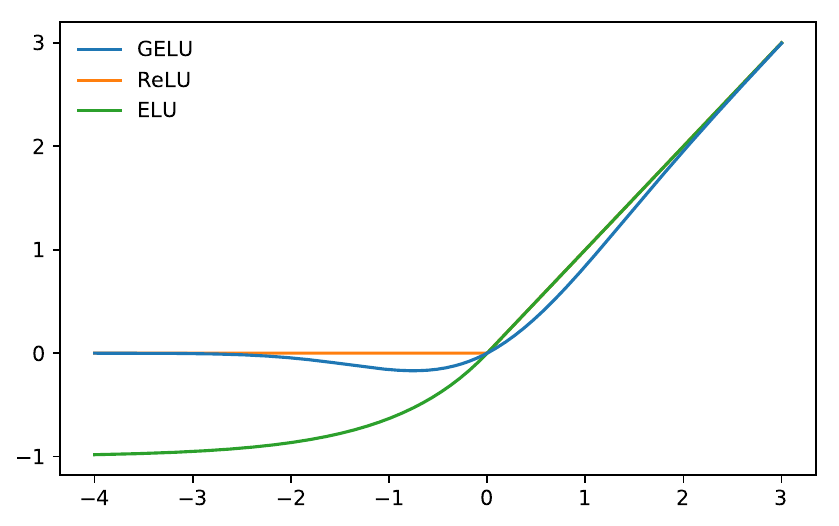}
  	\caption{The GELU ($\mu=0, \sigma=1$), ReLU, and ELU ($\alpha=1$).}
    \label{fig:nonlinearityplot}
\end{wrapfigure}
Masking inputs in this fashion retains non-determinism but maintains dependency upon the input value. A stochastically chosen mask amounts to a stochastic zero or identity transformation of the input. This is much like Adaptive Dropout \citep{standout}, but adaptive dropout is used in tandem with nonlinearities and uses a logistic not standard normal distribution. We found that it is possible to train competitive MNIST and TIMIT networks solely with this stochastic regularizer, all without using any nonlinearity.

We often want a deterministic decision from a neural network, and this gives rise to our new nonlinearity. The nonlinearity is the expected transformation of the stochastic regularizer on an input $x$, which is $\Phi(x)\times Ix + (1 - \Phi(x))\times 0x = x\Phi(x)$. Loosely, this expression states that we scale $x$ by how much greater it is than other inputs. Since the cumulative distribution function of a Gaussian is often computed with the error function, we define the Gaussian Error Linear Unit (GELU) as
\[
\text{GELU}(x) = xP(X\le x) = x\Phi(x) = x \cdot \frac{1}{2}\left[1 + \text{erf}(x/\sqrt{2})\right].
\]
We can approximate the GELU with
\[
0.5x (1 + \tanh[\sqrt{2/\pi}(x + 0.044715x^3)])
\]
or
\[
x \sigma(1.702 x),
\]
if greater feedforward speed is worth the cost of exactness.

We could use different CDFs. For example we could use Logistic Distribution CDF $\sigma(x)$ to get what we call the Sigmoid Linear Unit (SiLU) $x\sigma(x)$. We could use the CDF of $\mathcal{N}(\mu, \sigma^2)$ and have $\mu$ and $\sigma$ be learnable hyperparameters, but throughout this work we simply let $\mu = 0$ and $\sigma = 1$. Consequently, we do not introduce any new hyperparameters in the following experiments. In the next section, we show that the GELU  exceeds ReLUs and ELUs across numerous tasks.

\section{GELU Experiments}
We evaluate the GELU, ELU, and ReLU on MNIST classification (grayscale images with 10 classes, 60k training examples and 10k test examples), MNIST autoencoding, Tweet part-of-speech tagging (1000 training, 327 validation, and 500 testing tweets), TIMIT frame recognition (3696 training, 1152 validation, and 192 test audio sentences), and CIFAR-10/100 classification (color images with 10/100 classes, 50k training and 10k test examples). We do not evaluate nonlinearities like the LReLU because of its similarity to ReLUs (see \cite{lrelu} for a description of LReLUs).

\subsection{MNIST Classification}
\begin{figure}
	\centering
	\noindent\makebox[\textwidth]{\includegraphics[scale=0.38]{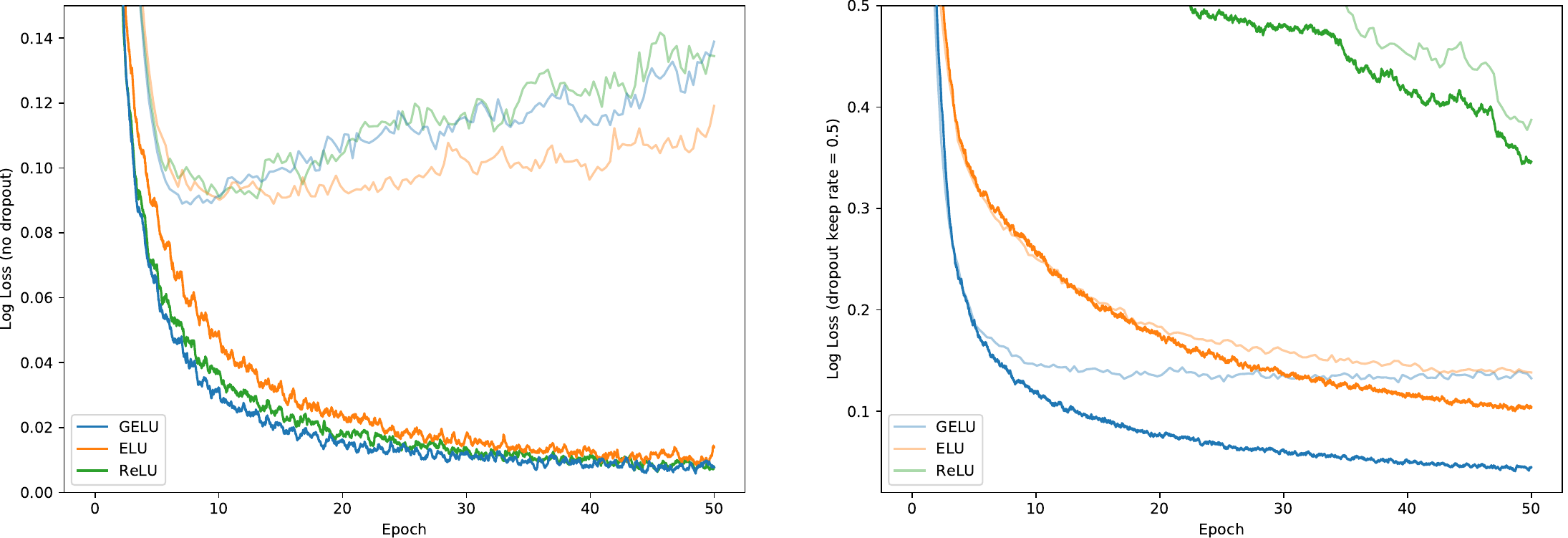}}
	\caption{MNIST Classification Results. Left are the loss curves without dropout, and right are curves with a dropout rate of $0.5$. Each curve is the the median of five runs. Training set log losses are the darker, lower curves, and the fainter, upper curves are the validation set log loss curves.}\label{fig:mnist1}
\end{figure}

\begin{figure}
	\centering
	\noindent\makebox[\textwidth]{\includegraphics[scale=0.5]{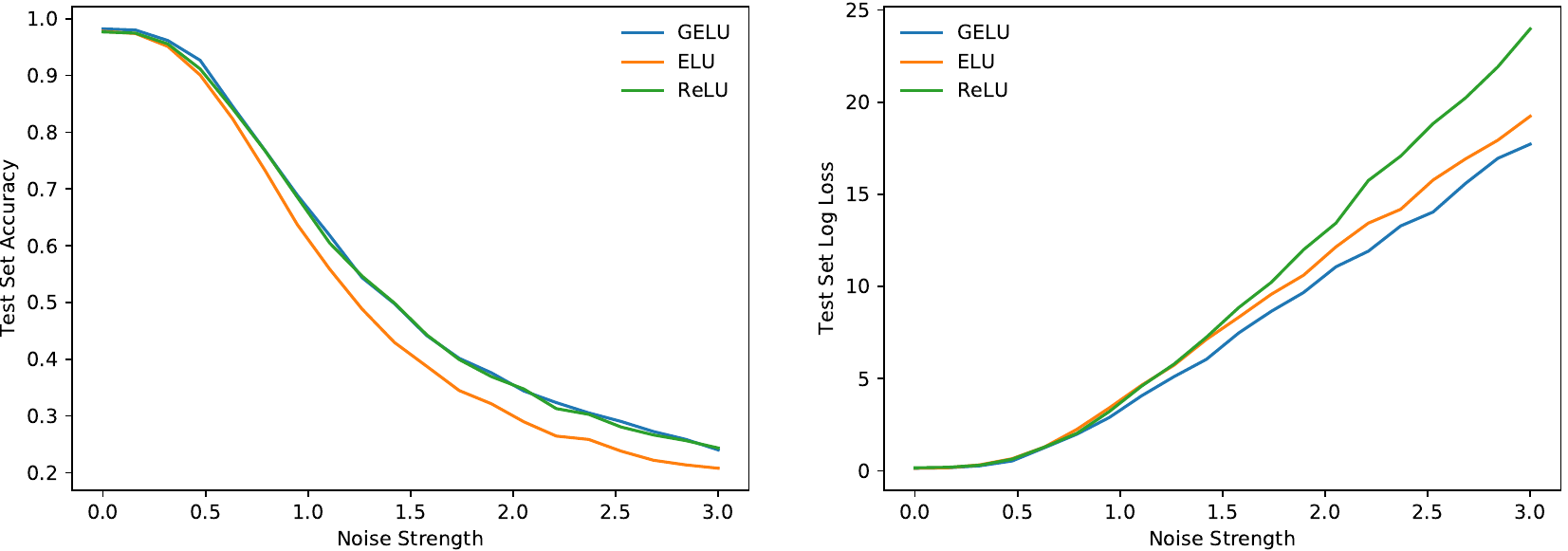}}
	\caption{MNIST Robustness Results. Using different nonlinearities, we record the test set accuracy decline and log loss increase as inputs are noised. The MNIST classifier trained without dropout received inputs with uniform noise $\text{Unif}[-a,a]$ added to each example at different levels $a$, where $a=3$ is the greatest noise strength. Here GELUs display robustness matching or exceeding ELUs and ReLUs.}\label{fig:robustness}
\end{figure}

Let us verify that this nonlinearity competes with previous activation functions by replicating an experiment from \citet{elu}. To this end, we train a fully connected neural network with GELUs ($\mu = 0, \sigma = 1$), ReLUs, and ELUs ($\alpha = 1$). Each 8-layer, 128 neuron wide neural network is trained for 50 epochs with a batch size of 128. This experiment differs from those of Clevert et al.~in that we use the Adam optimizer \citep{adam} rather than stochastic gradient descent without momentum, and we also show how well nonlinearities cope with dropout. Weights are initialized with unit norm rows, as this has positive impact on each nonlinearity's performance~\citep{hendrycks, weights, init2}. Note that we tune over the learning rates $\{10^{-3},10^{-4},10^{-5}\}$ with 5k validation examples from the training set and take the median results for five runs. Using these classifiers, we demonstrate in Figure \ref{fig:robustness} that classifiers using a GELU can be more robust to noised inputs. Figure~\ref{fig:mnist1} shows that the GELU tends to have the lowest median training log loss with and without dropout. Consequently, although the GELU is inspired by a different stochastic process, it comports well with dropout.

\subsection{MNIST Autoencoder}
\begin{figure}
	\centering
	\noindent\makebox[\textwidth]{\includegraphics[scale=0.38]{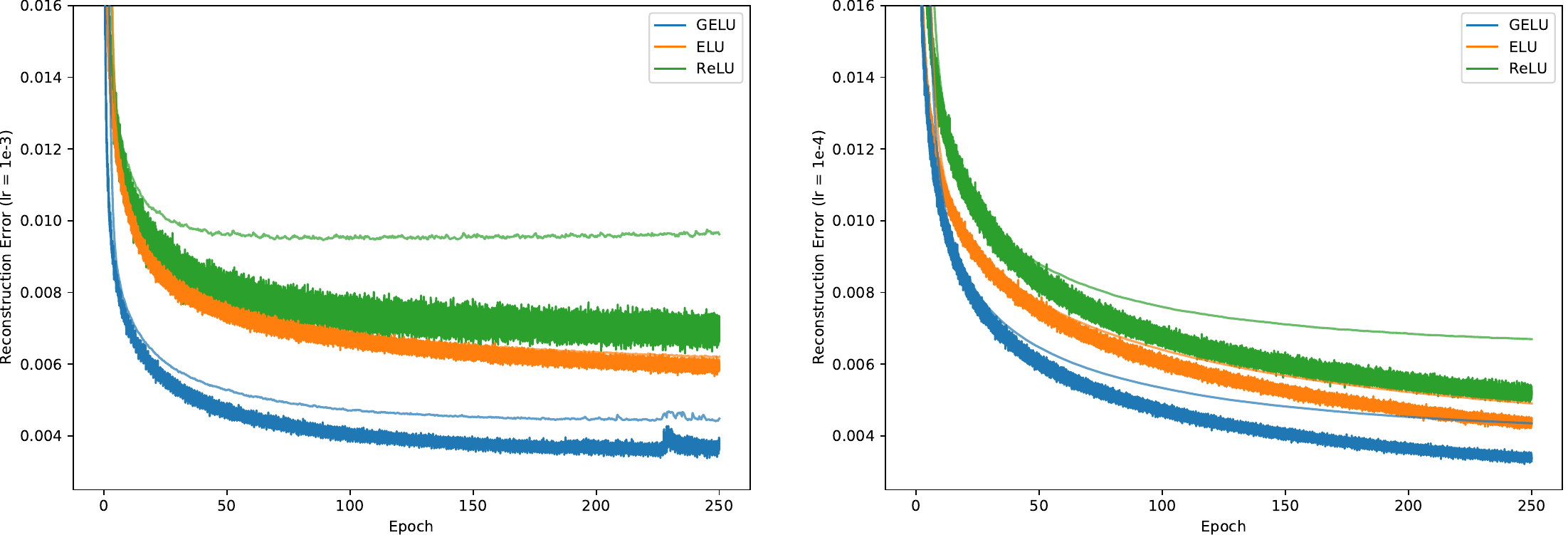}}
	\caption{MNIST Autoencoding Results. Each curve is the median of three runs. Left are loss curves for a learning rate of $10^{-3}$, and the 
right figure is for a $10^{-4}$ 
learning rate. Light, thin curves correspond to test set log losses.}\label{fig:mnist2}
\end{figure}

We now consider a self-supervised setting and train a deep autoencoder on MNIST \citep{desjardens}. To accomplish this, we use a network with layers of width 1000, 500, 250, 30, 250, 500, 1000, in that order. We again use the Adam optimizer and a batch size of 64. Our loss is the mean squared loss. We vary the learning rate from $10^{-3}$ to $10^{-4}$. We also tried a learning rate of $0.01$ but ELUs diverged, and GELUs and RELUs converged poorly. The results in Figure~\ref{fig:mnist2} indicate the GELU accommodates different learning rates and significantly outperforms the other nonlinearities.

\subsection{Twitter POS Tagging}
Many datasets in natural language processing are relatively small, so it is important that an activation generalize well from few examples. To meet this challenge we compare the nonlinearities on POS-annotated tweets \citep{gimpel,owoputi} which contain 25 tags. The tweet tagger is simply a two-layer network with pretrained word vectors trained on a corpus of 56 million tweets~\citep{owoputi}. 
The input is the concatenation of the vector of the word to be tagged and those of its left and right neighboring words. 
Each layer has 256 neurons, a dropout keep probability of 0.8, and the network is optimized with Adam while tuning over the learning rates $\{10^{-3}, 10^{-4}, 10^{-5}\}$. We train each network five times per learning rate, and the median test set error is 12.57\% for the GELU, 12.67\% for the ReLU, and 12.91\% for the ELU.

\subsection{TIMIT Frame Classification}

\begin{figure}
	\centering
	\noindent\makebox[\textwidth]{\includegraphics[scale=0.7]{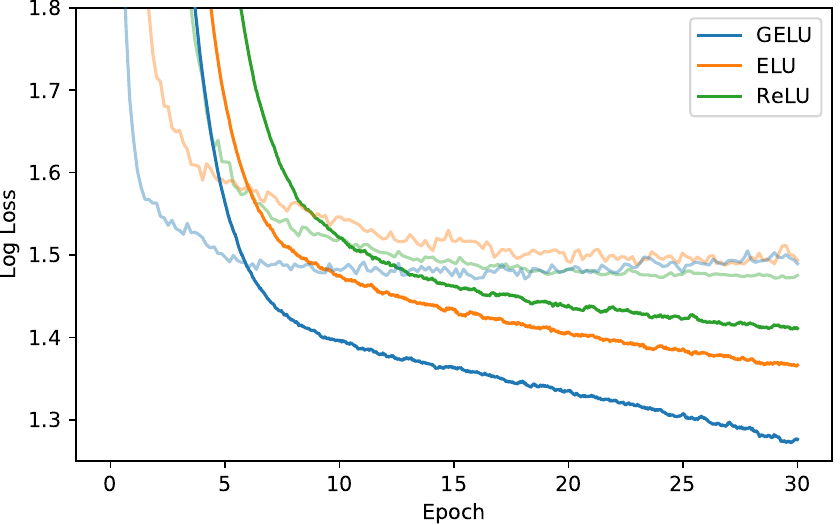}}
	\caption{TIMIT Frame Classification. Learning curves show training set convergence, and the lighter curves show the validation set convergence.}\label{fig:timit}
\end{figure}

Our next challenge is phone recognition with the TIMIT dataset which has recordings of 680 speakers in a noiseless environment. The system is a five-layer, 2048-neuron wide classifier as in~\citep{timitdeep} with 39 output phone labels and a dropout rate of 0.5 as in~\citep{srivastava}. This network takes as input 11 frames and must predict the phone of the center frame using 26 MFCC, energy, and derivative features per frame. We tune over the learning rates $\{10^{-3},10^{-4},10^{-5}\}$ and optimize with Adam. After five runs per setting, we obtain the median curves in Figure~\ref{fig:timit}, and median test error chosen at the lowest validation error is 29.3\% for the GELU, 29.5\% for the ReLU, and 29.6\% for the ELU.

\subsection{CIFAR-10/100 Classification}
Next, we demonstrate that for more intricate architectures the GELU nonlinearity again outperforms other nonlinearities. We evaluate this activation function using CIFAR-10 and CIFAR-100 datasets \citep{cifar} on shallow and deep convolutional neural networks, respectively.

\begin{figure}
	\vspace{-5pt}
	\centering
	\noindent\makebox[\textwidth]{\includegraphics[scale=0.7]{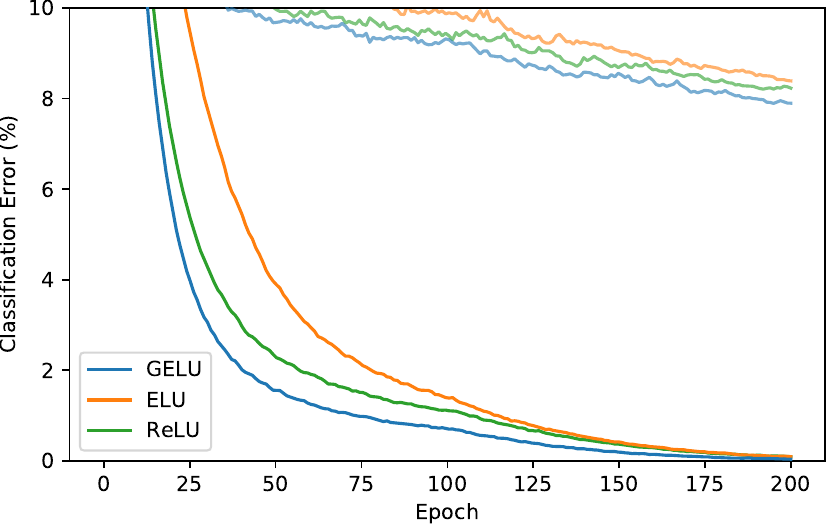}}
	\caption{CIFAR-10 Results. Each curve is the median of three runs. Learning curves show training set error rates, and the lighter curves show the test set error rates.}\label{fig:c10}
	\vspace{-5pt}
\end{figure}

Our shallower convolutional neural network is a 9-layer network with the architecture and training procedure from~\cite{weightnorm} while using batch normalization to speed up training. The architecture is described in \cref{appendixb} and recently obtained state of the art on CIFAR-10 without data augmentation. No data augmentation was used to train this network. We tune over the learning initial rates $\{10^{-3},10^{-4},10^{-5}\}$ with 5k validation examples then train on the whole training set again based upon the learning rate from cross validation. The network is optimized with Adam for 200 epochs, and at the 100th epoch the learning rate linearly decays to zero. Results are shown in Figure~\ref{fig:c10}, and each curve is a median of three runs. Ultimately, the GELU obtains a median error rate of \textbf{7.89}\%, the ReLU obtains 8.16\%, and the ELU obtains 8.41\%.

\begin{figure}
	\vspace{-5pt}
	\centering
	\noindent\makebox[\textwidth]{\includegraphics[scale=0.7]{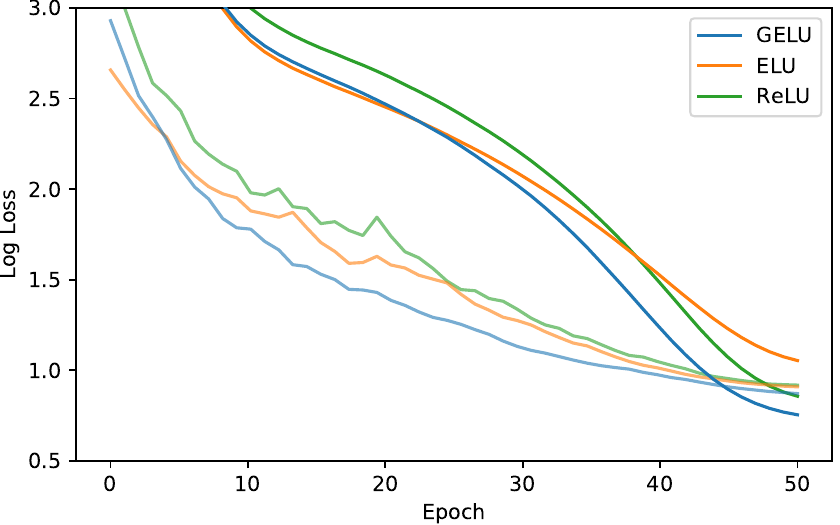}}
	\caption{CIFAR-100 Wide Residual Network Results. Learning curves show training set convergence with dropout on, and the lighter curves show the test set convergence with dropout off.}\label{fig:c100}
	\vspace{-5pt}
\end{figure}

Next we consider a wide residual network on CIFAR-100 with 40 layers and a widening factor of $4$ \citep{wrn}. We train for 50 epochs with the learning rate schedule described in~\citep{SGDR} ($T_0=50, \eta=0.1$) with Nesterov momentum, and with a dropout keep probability of 0.7. Some have noted that ELUs have an exploding gradient with residual networks \citep{eluresnet}, and this is alleviated with batch normalization at the end of a residual block. Consequently, we use a Conv-Activation-Conv-Activation-BatchNorm block architecture to be charitable to ELUs. Over three runs we obtain the median convergence curves in Figure~\ref{fig:c100}. Meanwhile, the GELU achieves a median error of \textbf{20.74}\%, the ReLU obtains 21.77\% (without our changes described above, the original 40-4 WideResNet with a ReLU obtains 22.89\% \citep{wrn}), and the ELU obtains 22.98\%.

\section{Discussion}
Across several experiments, the GELU outperformed previous nonlinearities, but it 
bears semblance to the ReLU and ELU in other respects. For example, as $\sigma \to 0$ and if $\mu = 0$, the GELU becomes a ReLU. More, the ReLU and GELU are equal asymptotically. In fact, the GELU can be viewed as a way to smooth a ReLU. To see this, recall that $\text{ReLU}=\max(x,0)=x\mathds{1}(x>0)$ (where $\mathds{1}$ is the indicator function), while the GELU is $x\Phi(x)$ if $\mu=0,\sigma=1$. Then the CDF is a smooth approximation to the binary function the ReLU uses, like how the sigmoid smoothed binary threshold activations. Unlike the ReLU, the GELU and ELU can be both negative and positive. In fact, if we used the cumulative distribution function of the standard Cauchy distribution, then the ELU (when $\alpha=1/\pi$) is asymptotically equal to $xP(C\le x), C \sim \textsf{Cauchy}(0,1)$ for negative values and for positive values is $xP(C\le x)$ if we shift the line down by $1/\pi$. These are some fundamental relations to previous nonlinearities.

However, the GELU has several notable differences. This non-convex, non-monotonic function is not linear in the positive domain and exhibits curvature at all points. Meanwhile ReLUs and ELUs, which are convex and monotonic activations, are linear in the positive domain and thereby can lack curvature. As such, increased curvature and non-monotonicity may allow GELUs to more easily approximate complicated functions than can ReLUs or ELUs. Also, since $\text{ReLU}(x)=x\mathds{1}(x>0)$ and $\text{GELU}(x)=x\Phi(x)$ if $\mu=0,\sigma=1$, we can see that the ReLU gates the input depending upon its sign, while the GELU weights its input depending upon how much greater it is than other inputs. In addition and significantly, the GELU has a probabilistic interpretation given that it is the expectation of a stochastic regularizer.



We also have two practical tips for using the GELU. First we advise using an optimizer with momentum when training with a GELU, as is standard for deep neural networks. Second, using a close approximation to the cumulative distribution function of a Gaussian distribution is important. A sigmoid function $\sigma(x)=1/(1+e^{-x})$ is an approximation of a cumulative distribution function of a normal distribution. However, we found that a Sigmoid Linear Unit (SiLU) $x\sigma(x)$ performs worse than GELUs but usually better than ReLUs and ELUs, so our SiLU is also a reasonable nonlinearity choice. Instead of using a $x\sigma(x)$ to approximate $\Phi(x)$, we used $0.5x (1 + \tanh[\sqrt{2/\pi}(x + 0.044715x^3)])$ \citep{approx}\footnote{Thank you to Dmytro Mishkin for bringing an approximation like this to our attention.} or $x \sigma(1.702 x)$. Both are sufficiently fast, easy-to-implement approximations, and we used the former in every experiment in this paper.

\section{Conclusion}
For the numerous datasets evaluated in this paper, the GELU exceeded the accuracy of the ELU and ReLU consistently, making it a viable alternative to previous nonlinearities.

\section*{Acknowledgment}
We would like to thank NVIDIA Corporation for donating several TITAN X GPUs used in this research.

\bibliographystyle{iclr2017_conference}
\bibliography{bibliography}

\newpage
\appendix

\section{Neural network architecture for CIFAR-10 experiments}
\label{appendixb}
\begin{table}[H]
\caption{Neural network architecture for CIFAR-10.}
\label{sample-table}
\begin{center}
\begin{tabular}{lll}
\multicolumn{1}{c}{\bf Layer Type}  &\multicolumn{1}{c}{\bf \# channels}  &\multicolumn{1}{c}{\bf $x,y$ dimension}
\\ \hline \\
raw RGB input         &3	&32\\
ZCA whitening         &3	&32\\
Gaussian noise $\sigma=0.15$         &3	&32\\
$3\times 3$ conv with activation         &96	&32\\
$3\times 3$ conv with activation         &96	&32\\
$3\times 3$ conv with activation         &96	&32\\
$2\times 2$ max pool, stride 2         &96	&16\\
dropout with $p=0.5$         &96	&16\\
$3\times 3$ conv with activation         &192	&16\\
$3\times 3$ conv with activation         &192	&16\\
$3\times 3$ conv with activation         &192	&16\\
$2\times 2$ max pool, stride 2         &192	&8\\
dropout with $p=0.5$         &192	&8\\
$3\times 3$ conv with activation         &192	&6\\
$1\times 1$ conv with activation         &192	&6\\
$1\times 1$ conv with activation         &192	&6\\
global average pool         			 &192	&1\\
softmax output		         			 &10	&1\\

\end{tabular}\label{tab:c10architecture}
\end{center}
\end{table}

\section{History of the GELU and SiLU}
This paper arose from DH’s first research internship as an undergraduate in June 2016. The start of the week after, this paper was put on arXiv, in which we discuss smoother ReLU activation functions ($x\times P(X\le x)$) and their relation to stochastic regularizers. In 2016, we submitted the paper to ICLR and made the paper and code publicly available. In the paper, we introduced and coined the Sigmoid Linear Unit (SiLU) as $x\cdot \sigma(x)$.

In the first half of 2017, Elfwing et al. published a paper that proposed the same activation function as SiLU, $x\cdot \sigma(x)$, which they called ``SIL.'' At the end of 2017, over a year after this paper was first released, Quoc Le and others from Google Brain put out a paper proposing $x\cdot \sigma(x)$ without citing either the Elfwing et al. paper or this work. Upon learning this, we contacted both parties. Elfwing quickly updated their work to call the activation the ``SiLU'' instead of ``SIL'' to recognize that we originally introduced the activation.

Unlike Elfwing et al., the Google Brain researchers continued calling the activation ``swish.'' However, there was no novelty. The first author of the ``swish'' paper stated their oversight in public, saying, ``As has been pointed out, we missed prior works that proposed the same activation function. The fault lies entirely with me for not conducting a thorough enough literature search.''  To subdue criticism, an update to the paper was released a week later. Rather than give credit to this work for the SiLU, the update only cited this work for the GELU so that the ``swish'' appeared more novel. In the updated paper, a learnable hyperparameter $\beta$ was introduced, and the swish was changed from $x\cdot \sigma(x)$ to $x\cdot \sigma(\beta \cdot x)$. This staked all of the idea's novelty on an added learnable hyperparameter $\beta$.

Despite the addition of the hyperparameter beta, nearly all of the community still used the original ``swish'' function without $\beta$ (i.e., with $\beta=1$). Since this paper was from Google Brain, the Tensorflow implementation ended up being called ``swish,'' and the default setting removed $\beta$, rendering it identical to the SiLU. The practice of adding an unused hyperparameter allowed claiming novelty while effectively receiving credit for an idea that originated elsewhere. Future papers with the same senior authors persistently referred to the ``swish'' function even when not using $\beta$, making it identical to the SiLU, originally proposed in this work. This resulted in the ``swish'' paper inappropriately gaining credit for the idea.

Things changed as the GELU began to be used in BERT and GPT, becoming the default activation for state-of-the-art Transformers. Now it is substantially more commonly used than the SiLU.

Separately, a reddit post ``Google has a credit assignment problem in research'' became popular and focused on how they refer to the SiLU as the swish. As an example, they mentioned ``Smooth Adversarial Training'' as an example of poor credit assignment. In the ``Smooth Adversarial Training'' paper, which came from the senior author of the swish, the term ``swish'' was used instead of ``SiLU.'' To reduce blowback from the post, the authors updated the paper and replaced ``swish'' with the ``SiLU,'' recognizing this paper as the original source of the idea. After this post, popular libraries such as Tensorflow and PyTorch also began to rename the function to ``SiLU'' instead of ``swish.'' For close observers, this issue has been largely settled, and we are grateful for the proper recognition that has largely come to pass.

\end{document}